# Identifying confounders using additive noise models

**Dominik Janzing**
MPI for Biol. Cybernetics
Spemannstr. 38
72076 Tübingen
Germany

**Jonas Peters**
MPI for Biol. Cybernetics
Spemannstr. 38
72076 Tübingen
Germany

**Joris Mooij**
MPI for Biol. Cybernetics
Spemannstr. 38
72076 Tübingen
Germany

**Bernhard Schölkopf**
MPI for Biol. Cybernetics
Spemannstr. 38
72076 Tübingen
Germany

## Abstract

We propose a method for inferring the existence of a latent common cause ("confounder") of two observed random variables. The method assumes that the two effects of the confounder are (possibly nonlinear) functions of the confounder plus independent, additive noise. We discuss under which conditions the model is identifiable (up to an arbitrary reparameterization of the confounder) from the joint distribution of the effects. We state and prove a theoretical result that provides evidence for the conjecture that the model is generically identifiable under suitable technical conditions. In addition, we propose a practical method to estimate the confounder from a finite i.i.d. sample of the effects and illustrate that the method works well on both simulated and real-world data.

## 1 Introduction

Since the pioneering work on causal inference methods (described for example in [Pearl, 2000] and [Spirtes et al., 1993]), much work has been done under the assumption that all relevant variables have been observed. An interesting, and possibly even more important, question is how to proceed if not all the relevant variables have been observed. In that case, dependencies between observed variables may also be explained by *confounders* — for instance, if a dependence between the incidence of storks and the birthrate is traced back to a common cause influencing both variables. In general, the difficulty not only lies in the fact that the values of the latent variables have not been observed, but also that the causal structure is unknown. In other words, it is in general not clear whether and how many confounders are needed to explain the data and which observed variables are directly caused by which confounder. Under the assumption of linear relationships between variables, confounders may be identified by means of Independent Component Analysis, as shown recently by Hoyer et al. [2008], if the distributions are non-Gaussian. Other results for the linear case are presented in Silva et al. [2006]. In this paper, we will not assume linear relationships, but try to tackle the more general, nonlinear case.

In the case of two variables without confounder, Hoyer et al. [2009] have argued that the causal inference task (surprisingly) becomes easier in case of nonlinear functional relationships. They have described a method to infer whether $X \to Y$ ("$X$ causes $Y$") or $Y \to X$ from the joint distribution $P(X, Y)$ of two real-valued random variables $X$ and $Y$. They consider models where $Y$ is a function $f$ of $X$ up to an additive noise term, i.e.,

$$Y = f(X) + N, \qquad (1)$$

where $N$ is an unobserved noise variable that is statistically independent of $X$. They show in their paper that generic choices of functions $f$, distributions of $X$ and distributions of $N$ induce joint distributions on $X, Y$ that do not admit such an additive noise model in the inverse direction, i.e., from $Y$ to $X$. If $P(X, Y)$ admits an additive model in one and only one direction, this direction can be interpreted as the true causal direction. We believe that the situation with a confounder between the two variables is similar in that respect: nonlinear functional relationships enlarge the class of models for which the causal structure is identifiable.

We now state explicitly which assumptions we make in the rest of this paper. First of all, we focus on the case of only two observed and one latent continuous random variables, all with values in $\mathbb{R}$. We assume that there is no feedback, or in other words, the true causal structure is described by a DAG (directed acyclic graph). Also, we assume that selection effects are absent, that is, the data samples are drawn i.i.d. from the probability distribution described by the model.

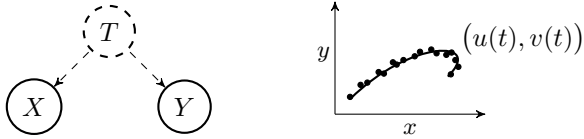

Figure 1: Directed acyclic graph and a scatter plot corresponding to a CAN model for two observed variables $X$ and $Y$ that are influenced by an unobserved confounder variable $T$.

**Definition 1** *Let $X$, $Y$ and $T$ be random variables taking values in $\mathbb{R}$. We define a model for* Confounders with Additive Noise (CAN) *by*

$$\begin{aligned} X &= u(T) + N_X \quad &\text{with } N_X, N_Y, T \\ Y &= v(T) + N_Y \quad &\text{jointly independent.} \end{aligned} \quad (2)$$

*where $u, v : \mathbb{R} \to \mathbb{R}$ are continuously differentiable functions and $N_X$, $N_Y$ are real-valued random variables.*

The random variables $N_X$ and $N_Y$ describe additive "noise", of which one may think of as the net effect of all other causes which are not shared by $X$ and $Y$. This model can be represented graphically by the DAG shown in Figure 1.

**Definition 2** *We call two CAN models equivalent if they induce the same distributions of $N_X$, $N_Y$ and the same joint distribution of $(u(T), v(T))$.*

This definition removes the ambiguity arising from unobservable reparameterizations of $T$. We further adopt the convention $\mathbb{E}(N_X) = \mathbb{E}(N_Y) = 0$.

The method we propose here enables one to distinguish between (i) $X \to Y$, (ii) $Y \to X$, and (iii) $X \leftarrow T \to Y$ for the class of models defined in (2), and (iv) to detect that no CAN model fits the data (which includes, for example, generic instances of the case that $X$ causes $Y$ and in addition $T$ confounds both $X$ and $Y$). If $N_X = 0$ a.s. ("almost surely") and $u$ is invertible, the model reduces to the model in (1) by setting $f := v \circ u^{-1}$. Given that we have observed a joint density on $\mathbb{R}^2$ of two variables $X, Y$ that admits a unique CAN model, the method we propose identifies this CAN model and therefore enables us to do causal inference by employing the following decision rule: we infer $X \to Y$ whenever $N_X$ is zero a.s. and $u$ invertible, infer $Y \to X$ whenever $N_Y$ is zero a.s. and $v$ invertible, and infer $X \leftarrow T \to Y$ if neither of the alternatives hold, which corresponds in spirit with Reichenbach's principle of common cause [Reichenbach, 1956]. Note that the case of $N_X = N_Y = 0$ a.s. and $u$ and $v$ invertible implies a deterministic relation between $X$ and $Y$, which we exclude here.

In practical applications, however, we propose to prefer the causal hypothesis $X \to Y$ already if the variance of $N_X$ is small compared to the variance of $N_Y$ (after we have normalized both $X$ and $Y$ to variance 1). To justify this, consider the case that $X$ causes $Y$ and the joint distribution admits a model (1), but by a slight measurement error, we observe $\tilde{X}$ instead of $X$, differing by a small additive noise term. Then $P(\tilde{X}, Y)$ admits a proper CAN model because $X$ is the latent common cause, but we infer $\tilde{X} \to Y$ because, from a coarse-grained point of view, we should not distinguish between the quantity $X$ and the measurement result $\tilde{X}$ if both variables almost coincide.

Finding the precise conditions under which the identification of CAN models is unique up to equivalence, is a non-trivial problem: If $u$ and $v$ are linear and $N_X, N_Y, T$ are Gaussian, one obtains a whole family of models inducing the same bivariate Gaussian joint distribution. Other examples where the model is not uniquely identifiable are given in Hoyer et al. [2009]: any joint density which admits additive noise models from $X$ to $Y$ and also from $Y$ to $X$ is a special case of a non-identifiable CAN model.

The remainder of this paper is organized as follows. In the next section, we provide theoretical motivation for our belief that in the generic case, CAN models are uniquely identifiable. A practical algorithm for the task is proposed in Section 3. It builds on a combination of nonlinear dimensionality reduction and kernel dependence measures. Section 4 provides empirical results on synthetic and real world data, and Section 5 concludes the paper.

## 2 Theoretical motivation

Below, we prove a partial identifiability result for the special case that both $u, v$ are invertible, where we consider the following limit: first, let the variances of the noise terms $N_X$ and $N_Y$ be small compared to the curvature of the graph $(u(t), v(t))$; second, we assume that the curvature is non-vanishing nevertheless (ruling out the linear case), and third, that the density on the graph $(u(t), v(t))$ changes slowly compared to the variance of the noise.

Because of the assumption that $u$ is invertible, we can reparameterize $T$ such that the CAN model (2) simplifies to

$$X = T + N_X \text{ and } Y = v(T) + N_Y\,,$$

i.e., the joint probability density is given by

$$p(x, y, t) = q(x-t)r(y-v(t))s(t)$$

where $q, r, s$ are the densities of $N_X$, $N_Y$ and $T$ respectively. We will further assume that $s$ is differentiable with bounded derivative and is nonzero at an infinite number of points. We also assume that the inverse function $w := v^{-1}$ of $v$ is two times differentiable and $w''$ is bounded. We further assume that the density of $N_X$ is several times differentiable, that all its moments are finite and that $\mathbb{E}(|N_X|^k)/k! \to 0$. This implies that the characteristic function of $N_X$, and therefore its distribution, can be uniquely expressed in terms of its moments. We assume likewise for $N_Y$.

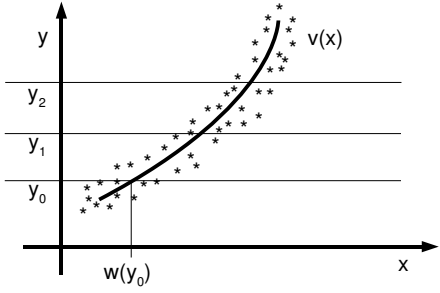

Figure 2: Sample from a distribution $p(x,y)$ obtained by adding small noise in $x$- and $y$-direction to a distribution supported by a graph $(t, v(t))$. We show that for fixed $y$, $X$ is approximately an affine function in the noise variables $N_X$ and $N_Y$.

We will show that under these assumptions, one can estimate the function $v$ from the conditional expectation $\mathbb{E}(X|Y=y)$ and that all moments of $N_X$ and $N_Y$ can be approximated from higher order conditional moments $\mathbb{E}(X^k|Y=y)$, with vanishing approximation error in the limit.

Defining $r_y(t) := p(y,t) = r(y - v(t))s(t)$, the conditional distribution of $T$ given $Y = y$ is given by $r_y(t)/p(y)$. For fixed $y$, we can locally (i.e., for $t \approx t_0 := w(y)$) approximate

$$r_y(t) \approx r\bigl(-v'(t_0)(t-t_0)\bigr)s(t_0)$$

if $s$ is almost constant around $t_0$ and if $v$ has small curvature at $t_0$. Hence,

$$T \mid y \sim w(y) + \beta_y N_Y \qquad (3)$$

approximately, where $\beta_y := -\frac{1}{v'(t_0)} = -w'(y)$.

We cannot observe $T$ directly, but we can observe the noisy version $X = T + N_X$ of it (see also Figure 2). From (3), we conclude that the conditional distribution of $X$ given $Y = y$ is approximately given by

$$X \mid y \sim w(y) + \beta_y N_Y + N_X \qquad (4)$$

Using $\mathbb{E}_y(\dots)$ as shorthand for the conditional expectation $\mathbb{E}(\dots \mid Y = y)$, we conclude that

$$\mathbb{E}_y(X) \approx w(y), \qquad (5)$$

because $N_X$ and $N_Y$ are centralized. We can thus approximately determine the function $v$ from observing all the conditional expectations $\mathbb{E}_y(X)$. After a shift by the estimated function $w(y)$, the conditional distribution of $X$, given $y$, is approximately given by a convolution of $N_X$ with the scaled noise $\beta_y N_Y$:

$$\bigl(X - \mathbb{E}_y(X)\bigr) \mid y \ \sim \ N_X + \beta_y N_Y$$

Since we can observe this convolution for different values $\beta_y$, we can compute the moments of $N_X$ and $N_Y$ using the following Lemma:

**Lemma 1** *Let $Z$ and $W$ be independent random variables for which all moments $\mathbb{E}(Z^k)$ and $\mathbb{E}(W^k)$ for $k = 1, \dots, n$ exist. Then all these moments can be reconstructed by observing the $n$th moments of $Z + c_j W$ for $n+1$ different values $c_0, \dots, c_n$.*

Proof: The $n$th moments of $Z + c_j W$ are given by

$$\mathbb{E}\bigl((Z+c_j W)^n\bigr) = \sum_{k=0}^{n} c_j^k \binom{n}{k} \mathbb{E}(Z^{n-k})\mathbb{E}(W^k).$$

The $(n+1) \times (n+1)$ matrix $M$ given by the entries $m_{jk} := c_j^k \binom{n}{k}$ is invertible because it coincides with the Vandermonde matrix up to multiplying the $k$th column with $\binom{n}{k}$. Hence we can compute all products $\mathbb{E}(Z^{n-k})\mathbb{E}(W^k)$ for $k = 0, \dots, n$ by matrix inversion and obtain in particular $\mathbb{E}(Z^n)$ and $\mathbb{E}(W^n)$. Taking into account only a subset of points, we can obviously compute lower moments, too. □

We now choose $n+1$ values $y_0, \dots, y_n$ such that we obtain $n+1$ different values $\beta_{y_0}, \dots, \beta_{y_n}$ and apply Lemma 1 to identify the first $n$ moments of $N_X$ and $N_Y$. For higher moments, the approximation will typically cause larger errors.

We now focus on the error made by the approximations above. We introduce the error terms

$$\epsilon_n(y) := \mathbb{E}_y\bigl((T - w(y))^n\bigr) - \mathbb{E}(\beta_y^n N_Y^n),$$

and hence obtain

$$\mathbb{E}_y(X) = \mathbb{E}_y(T) = w(y) + \epsilon_1(y).$$

Some calculations then yield the following exact relation between the moments of $N_X$ and $N_Y$:

$$\sum_{k=0}^{n} \beta_y^k \binom{n}{k} \mathbb{E}(N_X^{n-k})\mathbb{E}(N_Y^k) \qquad (6)$$

$$= \mathbb{E}_y((X - \mathbb{E}_y(X))^n) \qquad (7)$$

$$+ \sum_{k=1}^{n} \bigl(\epsilon_1(y)\bigr)^k \binom{n}{k} \mathbb{E}_y((X - \mathbb{E}_y(X))^{n-k}) \qquad (8)$$

$$- \sum_{k=0}^{n} \epsilon_k(y) \binom{n}{k} \mathbb{E}(N_X^{n-k}) \qquad (9)$$

Defining vectors **b**, **c** and **d** by letting their $j$'th component be the value of (7)–(9) for $y = y_j$, respectively, and defining the vector **q** and the matrix **M** by

$$q_k := \mathbb{E}(N_X^{n-k})\mathbb{E}(N_Y^k), \qquad M_{jk} := \beta_{y_j}^k \binom{n}{k},$$

we can write (6)–(9) for $y = y_0, \ldots, y_n$ in matrix notation as

$$\mathbf{Mq} = \mathbf{b} + \mathbf{c} + \mathbf{d}$$

and therefore

$$\mathbf{q} = \mathbf{M}^{-1}\mathbf{b} + \mathbf{M}^{-1}(\mathbf{c} + \mathbf{d}).$$

The above approximation yielded $\mathbf{q} = \mathbf{M}^{-1}\mathbf{b}$. The remaining terms are bounded from above by $\|\mathbf{M}^{-1}\|(\|\mathbf{c}\| + \|\mathbf{d}\|)$. The following lemma (which is proved in the Appendix) shows that the errors $\epsilon_k(y)$ are small. This then implies that the error in $\mathbf{q} \approx \mathbf{M}^{-1}\mathbf{b}$ is also of order $\mathcal{O}(\|s'\|_\infty + \|w''\|_\infty)$.

**Lemma 2** *For any $n$ and $y$ such that $s(w(y)) \neq 0$:*

$$\epsilon_n(y) = \mathcal{O}(\|s'\|_\infty + \|w''\|_\infty),$$

*where the terms involve moments of $N_Y$ and positive powers of $\beta_y$ and of $s(w(y))$, which are all bounded for given $y$ and $n$.*

We consider now a sequence of distributions $p_\ell(x, y)$ obtained by scaling the graph $(t, v(t))$ up to the larger graph $(\ell t, \ell v(t))$, while keeping the distributions of the noise $N_X$ and $N_Y$ fixed. Then we consider the conditional distributions of $X$, given $Y = y$ for $y = \ell y_0, \ldots, \ell y_n$ and can determine the moments of $N_X$ and $N_Y$ up to an error that converges to zero for $\ell \to \infty$, as the following Theorem shows.

**Theorem 1** *Define a sequence of joint densities $p_\ell(x, y)$ by*

$$p_\ell(x, y) := \int q(x - \ell t) r(y - \ell v(t)) s(t) dt. \qquad (10)$$

*Let, as above, $y_0, \ldots, y_n$ be chosen such that all $\beta_{y_0}, \ldots, \beta_{y_n}$ are different. Then every moment $\mathbb{E}(N_X^k)$ and $\mathbb{E}(N_Y^k)$ can be computed from the conditional moments $\mathbb{E}_{\ell y_j}(X^m)$ for $j = 0, \ldots, n$ and $m = 1, \ldots, n$ up to an error that vanishes for $\ell \to \infty$ under the assumptions made above.*

Proof: We rewrite eq. (10) as

$$p_\ell(x, y) := \int q(x - \tilde{t}_\ell) r(y - \tilde{v}_\ell(\tilde{t}_\ell)) \tilde{s}_\ell(\tilde{t}_\ell) d\tilde{t}_\ell.$$

with

$$\tilde{t}_\ell := \ell t, \quad \tilde{v}_\ell(\tilde{t}_\ell) := \ell v(\tilde{t}_\ell/\ell), \quad \tilde{s}_\ell(\tilde{t}_\ell) := s(\tilde{t}_\ell/\ell)/\ell.$$

$\tilde{w}_\ell$ is then defined as $\tilde{v}_\ell^{-1}$, and one checks easily that $\|\tilde{s}'_\ell\|_\infty$ and $\|\tilde{w}''_\ell\|_\infty$ tend to zero with $\mathcal{O}(1/\ell^2)$. Further, note that the $\beta_{y_j}$ are invariant with respect to the scaling: $\tilde{\beta}_{y_j,\ell} = \beta_{y_j}$. Hence, by Lemma 2, all $\epsilon_k(y_j)$ converge to zero. □

To summarize, we have sketched a proof that the CAN model becomes identifiable in a particular limit. We expect that a stronger statement holds (under suitable technical conditions), but postpone the non-trivial problem of finding the right technical conditions and the corresponding proof to future work.

The analysis above also shows that it should be possible to identify a confounder by estimating the variances of the noises $N_X$ and $N_Y$ and comparing their sizes (as discussed in the previous section). Indeed, in order to estimate the variance of the noise variable $N_X$, one observes the conditional expectation $\mathbb{E}(X \,|\, Y = y)$ for three different values of $y$; if the conditional expectations are sufficiently different, one can apply Lemma 1 to calculate the moments of $N_X$ and $N_Y$ up to second order. Assume, we observe a dependence between $X$ and $Y$ (which are normalized to have variance 1). In case one of the noise variances is much smaller than the other (say $N_X \ll N_Y$), this would indicate a direct causal influence ($X \to Y$ in that case), but if both noise variances are large, this indicates the existence of a confounder.

## 3  Identifying Confounders from Data

In this section we propose an algorithm (ICAN) that is able to identify a confounder in CAN models. While we only addressed the low noise regime in the previous theoretical section, the practical method we propose here should work even for strong noise, although in that case more data points are needed.

Assume that $X, Y$ are distributed according to the CAN model (2). We write $\mathbf{s}(t) := \big(u(t), v(t)\big)$ for the "true" curve of the confounder in $\mathbb{R}^2$. A scatter plot of the samples $(X, Y)$ (right panel of Figure 1, for example) suggests a simplistic method for detecting the confounder: for every curve $\mathbf{s} : [0, 1] \to \mathbb{R}^2$ project the data points $(X_k, Y_k)$ onto this curve $\mathbf{s}$, such that the Euclidean distance is minimized: $\hat{T}_k = \mathrm{argmin}_{t \in [0,1]} \|(X_k, Y_k) - \mathbf{s}(t)\|_2$. From a set of all possible paths $\mathcal{S}$ now choose the $\hat{\mathbf{s}}$ that minimizes the global $\ell_2$ distance $\sum_{k=1}^n \|(X_k, Y_k) - \mathbf{s}(\hat{T}_k)\|_2$ (dimensionality reduction) and propose $\hat{T}_k$ to be the confounder for $X_k$ and $Y_k$. This results in the estimated residuals $(\hat{N}_{X,k}, \hat{N}_{Y,k}) = (X_k, Y_k) - \hat{\mathbf{s}}(\hat{T}_k)$. If the hypotheses $\hat{T} \perp\!\!\!\perp \hat{N}_X, \hat{T} \perp\!\!\!\perp \hat{N}_Y, \hat{N}_X \perp\!\!\!\perp \hat{N}_Y$ cannot be rejected, propose that there is a confounder whose values are given by $\hat{T}_k$.

This idea turns out to be too naive: even if the data have been generated according to the model (2), the procedure results in dependent residuals. As an example, consider a data set simulated from the following model:

$$X = 4 \cdot \varphi_{-0.1}(T) + 4 \cdot \varphi_{1.1}(T) + N_X$$
$$Y = 1 \cdot \varphi_{-0.1}(T) - 1 \cdot \varphi_{1.1}(T) + N_Y$$

where $\varphi_\mu$ is the probability density of a $\mathcal{N}(\mu, 0.1^2)$ distributed random variable and $N_X, N_Y \sim U([-0.1, 0.1])$ and $T \sim U([0, 1])$ are jointly independent. We now minimize the global $\ell_2$ distance over the set of functions $\mathcal{S} = \{\mathbf{s} : \mathbf{s}_i(t) = \alpha_i \cdot \varphi_{-0.1}(t) + \beta_i \cdot \varphi_{1.1}(t); i = 1, 2\}$. Since there are only four parameters to fit, the problem is numerically solvable and gives the following optimal solution: $\alpha_1 = 3.9216, \beta_1 = 4.0112, \alpha_2 = 0.9776, \beta_2 = -0.9911$. The $\ell_2$ projections $\hat{T}$ result in a lower global $\ell_2$ distance (6.92) than the true values of $T$ (11.87).

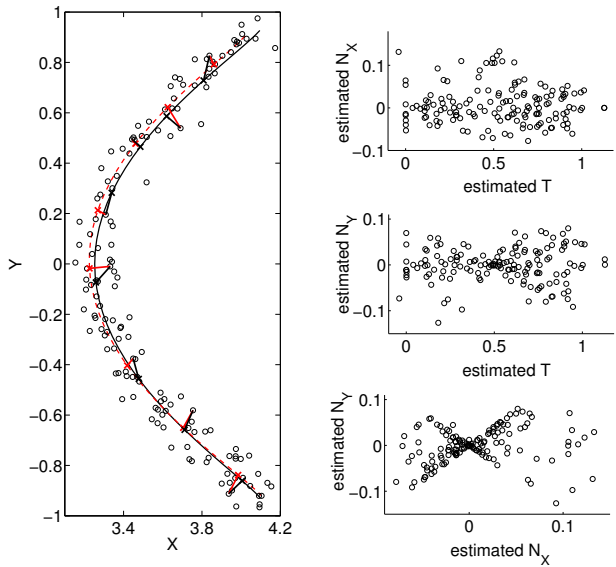

Figure 3: Left: a scatter plot of the data, true path $\mathbf{s}$ and projections (black and solid), estimated path $\hat{\mathbf{s}}$ and projections (red and dashed). Right: residuals plotted against each other and estimated confounder.

Figure 3 shows the true function $\mathbf{s}$ (black line), a scatter plot of $(X, Y)$ (black circles) and the computed curve $\hat{\mathbf{s}}$ that minimizes the global $\ell_2$ distance (dashed red line). Additionally, for some data points projections onto $\mathbf{s}$ and $\hat{\mathbf{s}}$ are shown, too: black crosses correspond to the "true projections" (i.e., the points without noise) onto $\mathbf{s}$ and red crosses correspond to projections onto the estimated function $\hat{\mathbf{s}}$ minimizing the $\ell_2$ distance. The latter result in the proposed residuals, which are shown together with the estimated values of the confounder on the right side of Figure 3. Estimated residuals and confounder values clearly depend on each other. Also independence tests like the Hilbert-Schmidt Independence Criterion (see below) reject the hypotheses of independence: $p$-values of $5 \times 10^{-2}$, $7 \times 10^{-5}$ and $1 \times 10^{-4}$ are computed corresponding to the right plots in Figure 3 from top to bottom. These dependencies often occur when the projections onto the curve $\hat{\mathbf{s}}$ are chosen to minimize the global $\ell_2$ distance, which can be seen as follows: in our example $\frac{\partial \mathbf{s}_1}{\partial t}$ is small for $T \approx 0.5$ or $Y \approx 0$. Since the points are projected onto the curve orthogonally, the projection results in very small residuals $\hat{N}_Y$. This introduces a dependency between $\hat{N}_Y$ and $\hat{T}$. Dependency between the residuals $\hat{N}_Y$ and $\hat{N}_X$ can arise from regions, where $\hat{\mathbf{s}}$ is approximately linear, like in the bottom right part of Figure 3: positive residuals $N_Y$ lead to positive residuals $N_X$ and vice versa.

Summarizing, projecting the pairs $(X, Y)$ onto the path $(\hat{\mathbf{s}}(t))$ by minimizing the $\ell_2$ distance to the path is the wrong approach for our purpose. Instead, the data $(X, Y)$ should be projected in a way that minimizes the dependence between residuals and confounder ($\hat{N}_X, \hat{T}$ and $\hat{N}_Y, \hat{T}$) and between the residuals itself ($\hat{N}_X, \hat{N}_Y$).

Let DEP($W, Z$) denote any non-negative dependence measure between random variables $W$ and $Z$, which is zero if and only if $W$ and $Z$ are independent (we later suggest to use the Hilbert-Schmidt Independence Criterion). In the example above we can use the red curve as an initial guess, but choosing the projections by minimizing the sum of the three dependence measures instead of $\ell_2$ distances. In our example this indeed leads to residuals that fulfill the independence constraints (p-values of 1.00, 0.65, 0.80). For the general case, we propose Algorithm 1 as a method for identifying the hidden confounder $T$ given an i.i.d. sample of $(X, Y)$.

If a CAN model can be found we interpret the outcome of our algorithm as $X \to Y$ if $\frac{\mathbb{V}\mathrm{ar}\hat{N}_X}{\mathbb{V}\mathrm{ar}\hat{N}_Y} \ll 1$ and $\hat{u}$ invertible and as $Y \to X$ if $\frac{\mathbb{V}\mathrm{ar}\hat{N}_X}{\mathbb{V}\mathrm{ar}\hat{N}_Y} \gg 1$ and $\hat{v}$ invertible. There is no mathematical rule that tells whether one should identify a variable $X$ and its (possibly noisy) measurement $\tilde{X}$ or consider them as separate variables instead. Thus we cannot be more explicit about the threshold for the factor between $\mathbb{V}\mathrm{ar}\hat{N}_X$ and $\mathbb{V}\mathrm{ar}\hat{N}_Y$ that tells us when to accept $X \to Y$ or $Y \to X$ or $X \leftarrow T \to Y$.

To implement the method we still need an algorithm for the initial dimensionality reduction, a dependence criterion DEP, a way to minimize it and an algorithm for non-linear regression. Surely, many different choices are possible. We will now briefly justify our choices for the implementation.

**Algorithm 1** Identifying Confounders using Additive Noise Models (ICAN)

1: **Input:** $(X_1, Y_1), \ldots, (X_n, Y_n)$ (normalized)
2: **Initialization:**
3: Fit a curve $\hat{\mathbf{s}}$ to the data that minimizes $\ell_2$ distance: $\hat{\mathbf{s}} := \operatorname{argmin}_{\mathbf{s} \in \mathcal{S}} \sum_{k=1}^n \operatorname{dist}(\mathbf{s}, (X_k, Y_k))$.
4: **repeat**
5:   **Projection:**
6:   $\hat{T} := \operatorname{argmin}_T \operatorname{DEP}(\hat{N}_X, \hat{N}_Y) + \operatorname{DEP}(\hat{N}_X, T) + \operatorname{DEP}(\hat{N}_Y, T)$ with $(\hat{N}_{X,k}, \hat{N}_{Y,k}) = (X_k, Y_k) - \hat{\mathbf{s}}(T_k)$
7:   **if** $\hat{N}_X \perp\!\!\!\perp \hat{N}_Y$ and $\hat{N}_X \perp\!\!\!\perp \hat{T}$ and $\hat{N}_Y \perp\!\!\!\perp \hat{T}$ **then**
8:     **Output:** $(\hat{T}_1, \ldots, \hat{T}_n)$, $\hat{u} = \hat{\mathbf{s}}_1$, $\hat{v} = \hat{\mathbf{s}}_2$, and $\frac{\mathbb{V}\mathrm{ar}\hat{N}_X}{\mathbb{V}\mathrm{ar}\hat{N}_Y}$.
9:     **Break.**
10:   **end if**
11:   **Regression:**
12:   Estimate $\hat{\mathbf{s}}$ by regression $(X, Y) = \hat{\mathbf{s}}(\hat{T}) + \hat{\mathbf{N}}$. Set $\hat{u} = \hat{\mathbf{s}}_1, \hat{v} = \hat{\mathbf{s}}_2$.
13: **until** $K$ iterations
14: **Output:** Data cannot be fitted by a CAN model.

**Initial Dimensionality Reduction**
It is difficult to solve the optimization problem (line 3 of the algorithm) for a big function class $\mathcal{S}$. Our approach thus separates the problem into two parts: we start with an initial guess for the projection values $\hat{T}_k$ (this is chosen using an implementation of the Isomap algorithm [Tenenbaum et al., 2000] by van der Maaten [2007]) and then iterate between two steps: In the first step we keep the projection values $\hat{T}_k$ fixed and choose a new function $\hat{\mathbf{s}} = (\hat{u}, \hat{v})$, where $\hat{u}$ and $\hat{v}$ are chosen by regression from $X$ on $\hat{T}$ and $Y$ on $\hat{T}$, respectively. To this end we used Gaussian Process Regression [Rasmussen and Williams, 2006], using the implementation of Rasmussen and Williams [2007], with hyperparameters set by maximizing marginal likelihood. In the second step the curve is fixed and each data point $(X_k, Y_k)$ is projected to the nearest point of the curve: $T_k$ is chosen such that $\|\hat{\mathbf{s}}(T_k) - (X_k, Y_k)\|_{\ell_2}$ is minimized. We then perform the first step again. A similar iterative procedure for dimensionality reduction has been proposed by Hastie and Stuetzle [1989].

This initial step of the algorithm is used for stabilization. Although the true curve $\mathbf{s}$ may differ from the $\ell_2$ minimizer $\hat{\mathbf{s}}$, the difference is not expected to be very large. Minimizing dependence criteria from the beginning often results in very bad fits.

**Dependence Criterion and its Minimization**
There are various choices of dependence criteria that can be used for the algorithm. Notice, however, that they should be able both to deal with continuous data and to detect non-linear dependencies. Since there is no canonical way of discretizing continuous variables, methods that work for discrete data (like a $\chi^2$ test) are not suitable for our purpose. In our method we chose the Hilbert-Schmidt Independence Criterion (HSIC) [Gretton et al., 2005, 2008]. It can be defined as the distance between the joint distribution and the product of the marginal distribution represented in a Reproducing Kernel Hilbert Space. For specific choices of the kernel (e.g., a Gaussian kernel) it has been shown that HSIC is zero if and only if the two distributions are independent. Furthermore the distribution of HSIC under the hypothesis of independence can be approximated by a Gamma distribution [Kankainen, 1995]. Thus we can construct a statistical test for the null hypothesis of independence. In our experiments we used Gaussian kernels and chose their kernel sizes to be the median distances between the points [Schölkopf and Smola, 2002]: e.g. $2\sigma^2 = \operatorname{median}\{\|X_k - X_l\|^2 : k < l\}$. We will use the term HSIC for the value of the Hilbert-Schmidt norm and $p_{\mathrm{HSIC}}$ for the corresponding $p$-value. For a small $p$-value ($< 0.05$, say) the hypothesis of independence is rejected.

For the projection step we now minimize $\operatorname{HSIC}(\hat{N}_X, \hat{N}_Y) + \operatorname{HSIC}(\hat{N}_X, \hat{T}) + \operatorname{HSIC}(\hat{N}_Y, \hat{T})$ with respect to $\hat{T}$. Note that at this part of the algorithm the function $\hat{\mathbf{s}}$ (and thus $\hat{u}$ and $\hat{v}$) remain fixed and the residuals are computed according to $N_X = X - \hat{u}(\hat{T})$ and $N_Y = Y - \hat{v}(\hat{T})$. We used a standard optimization algorithm for this task (`fminsearch` in MatLab) initializing it with the values of $\hat{T}$ obtained in the previous iteration. Instead of the sum of the three dependence criterion the maximum can be used, too. This is theoretically possible, but complicates the optimization problem since it introduces non-differentiability.

It should be mentioned that sometimes (not for all data sets though) a regularization for the $T$ values may be needed. Even for dependent noise very positive (or negative) values of $T$ result in large residuals, which may be regarded as independent. In our implementation we used a heuristic and just performed 5000 iterations of the minimization algorithm, which proved to work well in practice.

**Non-linear Regression**
Here, again, we used Gaussian process regression for both variables separately. Since the confounder values $\hat{T}$ are fixed we can fit $X = \hat{u}(\hat{T}) + \hat{N}_X$ and $Y = \hat{v}(\hat{T}) + \hat{N}_Y$ to obtain $\hat{\mathbf{s}} = (\hat{u}, \hat{v})$.

In the experiments this step was mostly not necessary: whenever the algorithm was able to find a solution with independent residuals, it did so in the first or second iteration after optimizing the projections according to the dependence measures. We still think that this step can be useful for difficult data sets, where the curve that minimizes the $\ell_2$ distance is very different from the ground truth.

## 4 Experiments

In this section we show that our method is able to detect confounders both in simulated and in real data.

### 4.1 Simulated data

**Data set 1.**
We show on a simulated data set that our algorithm finds the confounder if the data comes from the model assumed in (2). We simulated 200 data points from a curve whose components $u$ and $v$ consist of a random linear combination of Gaussian bumps each. The noise is uniformly distributed on $[-0.035, 0.035]$. Note that in contrast to the example given in Section 3 we are now doing the regression using Gaussian processes. The algorithm finds a curve and corresponding projections of the data points onto this curve, such that $\hat{N}_X, \hat{N}_Y$ and $\hat{T}$ are pairwise independent, which can be seen from the $p$-values $p_{\text{HSIC}}(\hat{N}_X, \hat{N}_Y) = 0.94$, $p_{\text{HSIC}}(\hat{N}_X, \hat{T}) = 0.78$ and $p_{\text{HSIC}}(\hat{N}_Y, \hat{T}) = 0.23$. The top panel of Figure 4 shows the data and both true (black) and estimated (red) curve. The bottom panel shows estimated and true values of the confounder. Recall that the confounder can be estimated only up to an arbitrary reparameterization (e.g. $t \mapsto -t$).

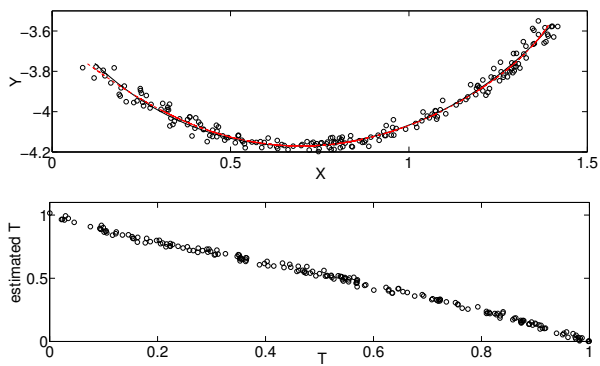

Figure 4: Data set 1. Top: true (black) and estimated (red) curve. Bottom: The estimated values of the confounder are plotted against the true values. Apart from the arbitrary reparameterization $t \mapsto -t$ the method inferred confounder values close to the true ones.

In this example the empirical joint distribution of $(X, Y)$ does not allow a simple direct causal relationship between $X$ and $Y$. It is obvious that the data cannot be explained by $X = g(Y) + N$ with a noise $N$ that is independent of $Y$. It turns out that also the model corresponding to the other direction $X \to Y$ can be rejected since a regression of $Y$ onto $X$ leads to dependent residuals ($p_{\text{HSIC}}(X, Y - \hat{f}(X)) = 0.0015$).

**Data set 2.**
This data set is produced in the same way as data set 1, but this time using an invertible $v$ and unequal scaled noises. We sampled $N_X \sim U([-0.008, 0.008])$ and $N_Y \sim U([-0.0015, 0])$. We argued above that for finite sample sizes this case should rather be regarded as $Y \to X$ and not as an example with a hidden common cause. The algorithm again identifies a curve and projections, such that the independence constraints are satisfied ($p_{\text{HSIC}}(\hat{N}_X, \hat{N}_Y) = 1.00$, $p_{\text{HSIC}}(\hat{N}_X, \hat{T}) = 1.00$ and $p_{\text{HSIC}}(\hat{N}_Y, \hat{T}) = 1.00$, see Figure 5), and it is important to note that the different scales of the variances are reduced, but still noticeable ($\frac{\mathbb{V}\text{ar}(\hat{N}_X)}{\mathbb{V}\text{ar}(\hat{N}_Y)} \approx 5$). In such a case we indeed interpret the outcome of our algorithm as "$Y$ causes $X$".

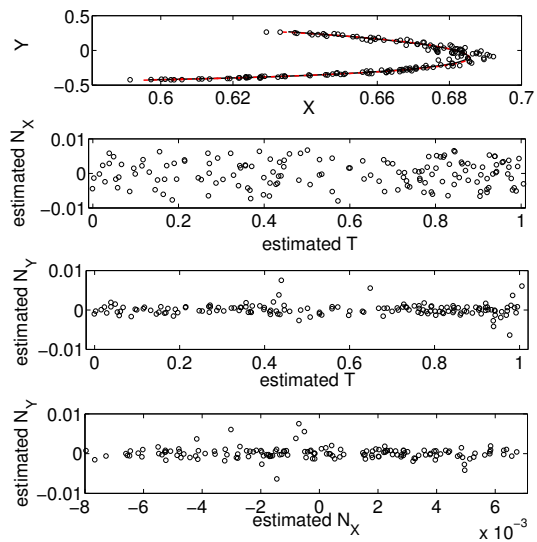

Figure 5: Data set 2. Top: true (black) and estimated (red) curve. Others: Scatter plots of the fitted residuals against each other and against estimated values for the confounder. The fact that the noise $N_X$ has been sampled with a higher variance than $N_Y$ can also be detected in the fitted residuals.

Since the variances of $N_X$ and $N_Y$ differ significantly and the sample size is small, we can (as expected) even fit a direct causal relationship between $X$ and $Y$: Assuming the model

$$X = g(Y) + N \quad (11)$$

and fitting the function $\hat{g}$ by Gaussian Process re-

gression, for example, results in independent residuals: $p_{\text{HSIC}}(Y, X - \hat{g}(Y)) = 0.97$. Thus we regard the model (11) and thus $Y \to X$ to be true. This does not contradict the identifiability conjecture because the dependencies introduced by setting the noise $\hat{N}_Y$ mistakenly to zero are not detectable at this sample size.

**Data set 3.**

We also simulated a data set for which the noise terms $N_X$ and $N_Y$ clearly depend on $T$. Figure 6 shows a scatter plot of the data set, the outcome curve of the algorithm after $K = 5$ iterations (top) and a scatter plot between the estimated residuals $\hat{N}_Y$ and confounder values $\hat{T}$ (bottom). The method did not find a curve and corresponding projections for which the residuals were independent ($p_{\text{HSIC}}(\hat{N}_Y, \hat{T}) = 0.00$, for example), and thus results in "Data cannot be fitted by a CAN model". This makes sense since the data set was not simulated according to model (2).

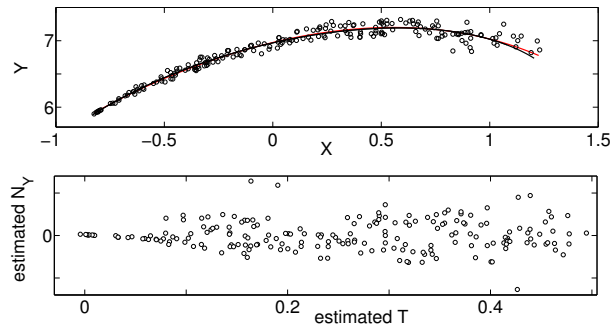

Figure 6: Data set 3. To check whether our method does not always find a confounder we simulated a data set where the noise clearly depends on $T$. Indeed the algorithm does not find an independent solution and stops after $K = 5$ iterations. Top: true (black) and estimated (red) curve. Bottom: the estimated residuals clearly depend on the estimated confounder.

### 4.2 Real data

**ASOS data.**

The Automated Surface Observations Systems (ASOS) consists of several stations that automatically collect and transmit weather data every minute. We used 150 values for air pressure that were collected by stations KABE and KABI in January 2000 [NCDC, 2009]. We expect the time to be a confounder. As in the other experiments a projection minimizing the $\ell_2$ distance would not be sufficient: after the initialization step we obtain $p$-values, which reject independence ($p_{\text{HSIC}}(\hat{N}_X, \hat{N}_Y) = 0.00$, $p_{\text{HSIC}}(\hat{N}_X, \hat{T}) = 0.00$, $p_{\text{HSIC}}(\hat{N}_Y, \hat{T}) = 0.02$). After the projection step minimizing the sum of HSICs the residuals are regarded as independent: $p_{\text{HSIC}}(\hat{N}_X, \hat{N}_Y) = 1.00$, $p_{\text{HSIC}}(\hat{N}_X, \hat{T}) = 1.00$, $p_{\text{HSIC}}(\hat{N}_Y, \hat{T}) = 0.16$. Figures 7 and 8 show the results. The confounder has been successfully identified.

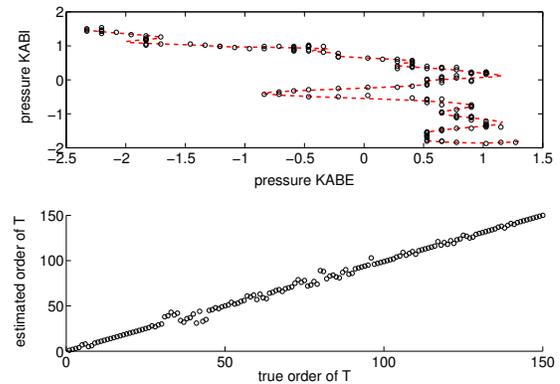

Figure 7: ASOS data. Top: scatter plot of the data, together with the estimated path $\hat{\mathbf{s}}$ (note that it is not interpolating between the data points). Bottom: ordering of the estimated confounder values against the true ordering. The true ordering is almost completely recovered.

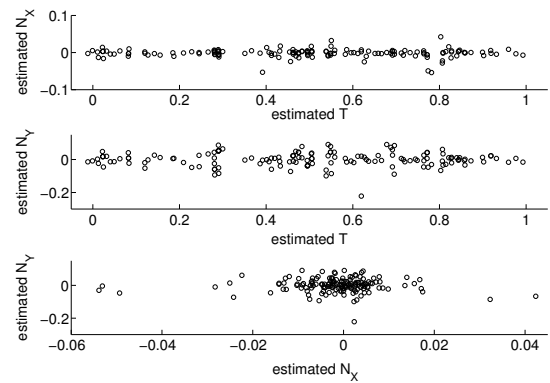

Figure 8: ASOS data. Residuals plotted against each other and against the estimated confounder. The hypothesis of independence is not rejected, which means the method identified the confounder.

## 5 Conclusion and Future Work

We have proposed a method to identify the confounder of a model where two observable variables are functions of an unobserved confounder plus additive noise. We have provided a theoretical motivation for the method, and showed that the algorithm works on both simulated and real world data sets.

Our initial results are encouraging, and our theoretical motivation provides some insight into why the problem is solvable in the first place.

A complete identifiability result in the style of Hoyer et al. [2009], however, would clearly be desirable, along with further experimental evidence.

## Appendix: proof of Lemma 2

Fix $y$ and set $t_y := w(y)$, $s_y := s(w(y))$, $\beta_y := -w'(y)$. We compute:

$$p(y)\mathbb{E}_y((T-t_y)^n) = \int (t-t_y)^n r(y-v(t))s(t)\,dt.$$

We want to make the substitution $\tilde{y} = y - v(t)$, i.e., $t = w(y - \tilde{y})$. Application of Taylor's theorem yields:

$$\begin{aligned}w(y-\tilde{y}) - t_y &= \beta_y \tilde{y} + \tilde{y}^2 \epsilon & |\epsilon| &\leq \|w''\|_\infty/2, \\ -w'(y-\tilde{y}) &= \beta_y + \eta \tilde{y} & |\eta| &\leq \|w''\|_\infty, \\ s(w(y-\tilde{y})) &= s_y + \zeta(\beta_y \tilde{y} + \tilde{y}^2 \epsilon) & |\zeta| &\leq \|s'\|_\infty.\end{aligned}$$

where we suppress the dependencies of $\epsilon, \eta, \zeta$ on $\tilde{y}$ in the notation. Therefore:

$$\int (t-t_y)^n r(y-v(t))s(t)\,dt$$
$$= \int (\beta_y \tilde{y} + \tilde{y}^2 \epsilon)^n r(\tilde{y})\bigl(s_y + \zeta(\beta_y \tilde{y} + \tilde{y}^2 \epsilon)\bigr)(\beta_y + \eta \tilde{y})\,d\tilde{y}.$$

The special case $n = 0$ yields

$$p(y) = \int r(\tilde{y})\bigl(s_y + \zeta(\beta_y \tilde{y} + \tilde{y}^2 \epsilon)\bigr)(\beta_y + \eta \tilde{y})\,d\tilde{y}$$
$$= s_y \beta_y + \mathcal{O}(\|s'\|_\infty + \|w''\|_\infty).$$

For arbitrary $n$, we obtain

$$\int (\beta_y \tilde{y} + \tilde{y}^2 \epsilon)^n r(\tilde{y})\bigl(s_y + \zeta(\beta_y \tilde{y} + \tilde{y}^2 \epsilon)\bigr)(\beta_y + \eta \tilde{y})\,d\tilde{y}$$
$$= s_y \beta_y \mathbb{E}\bigl((\beta_y N_Y)^n\bigr) + \mathcal{O}(\|s'\|_\infty + \|w''\|_\infty).$$

The error terms contain moments of $N_Y$, which are all finite by assumption, and (positive) powers of $s_y$ and $\beta_y$, which are also finite. Thus we conclude that

$$\epsilon_n(y) = \mathbb{E}_y((T-t_y)^n) - \mathbb{E}(\beta_y^n N_Y^n) = \mathcal{O}(\|s'\|_\infty + \|w''\|_\infty).$$

　□